\documentclass[letterpaper, 10 pt, conference]{ieeeconf} 

\usepackage{cite}
\usepackage{amsmath,amssymb,amsfonts}
\usepackage{algorithmic}
\usepackage{graphicx}
\usepackage{textcomp}
\usepackage{pbalance}
\usepackage{xcolor}
\usepackage{hyperref}

\title{Quadrupedal Robot Skateboard Mounting via Reverse Curriculum Learning}

\newcommand{\IEEEauthorblockN}[1]{\textbf{#1}}
\newcommand{\IEEEauthorblockA}[1]{\textit{#1}}

\author{
\begin{tabular}{ccc}
\IEEEauthorblockN{1\textsuperscript{st} Danil Belov} &
\IEEEauthorblockN{2\textsuperscript{nd} Artem Erkhov} &
\IEEEauthorblockN{3\textsuperscript{rd} Elizaveta Pestova} \\
\IEEEauthorblockA{\textit{Skolkovo Inst. of Sci. and Tech.}} &
\IEEEauthorblockA{\textit{Skolkovo Inst. of Sci. and Tech.}} &
\IEEEauthorblockA{\textit{Skolkovo Inst. of Sci. and Tech.}} \\
Moscow, Russia &
Moscow, Russia &
Moscow, Russia \\
Danil.Belov@skoltech.ru &
Artem.Erkhov@skoltech.ru &
Elizaveta.Pestova@skoltech.ru \\
[12pt]
\IEEEauthorblockN{4\textsuperscript{th} Ilya Osokin} &
\IEEEauthorblockN{5\textsuperscript{th} Dzmitry Tsetserukou} &
\IEEEauthorblockN{6\textsuperscript{th} Pavel Osinenko} \\
\IEEEauthorblockA{\textit{Skolkovo Inst. of Sci. and Tech.}} &
\IEEEauthorblockA{\textit{Skolkovo Inst. of Sci. and Tech.}} &
\IEEEauthorblockA{\textit{Skolkovo Inst. of Sci. and Tech.}} \\
Moscow, Russia &
Moscow, Russia &
Moscow, Russia \\
Ilya.Osokin@skoltech.ru &
D.Tsetserukou@skoltech.ru &
Pavel.Osinenko@skoltech.ru
\end{tabular}
}

\begin{document}

\maketitle

\begin{abstract}

The aim of this work is to enable quadrupedal robots to mount skateboards using Reverse Curriculum Reinforcement Learning.
Although prior work has demonstrated skateboarding for quadrupeds that are already positioned on the board, the initial mounting phase still poses a significant challenge.
A goal-oriented methodology was adopted, beginning with the terminal phases of the task and progressively increasing the complexity of the problem definition to approximate the desired objective.
The learning process was initiated with the skateboard rigidly fixed within the global coordinate frame and the robot positioned directly above it. Through gradual relaxation of these initial conditions, the learned policy demonstrated robustness to variations in skateboard position and orientation, ultimately exhibiting a successful transfer to scenarios involving a mobile skateboard.

The code, trained models, and reproducible examples are available at the following link:
https://github.com/dancher00/quadruped-skateboard-mounting




\end{abstract}

\section{Introduction}

Legged robot locomotion has a number of advantages over the other motion types.
The main one is the versatility, that comes from the ability of the legs to change their contact points.
Legged locomotion is naturally omnidirectional, and allows for the obstacles and difficult terrain to be traversed.

On the other hand, it is extremely difficult on many levels.
The mechanical construction of the robot should be able to carry the weight of the robot in the presence of sudden impacts.
The motors are supposed to have both high velocity and high torque in order to perform walking and running.

Legged locomotion took several decades to develop to its current state.
The first electrically and hydraulically actuated legged machines were proposed as early as in the 1950s.
At that time they were heavy and slow.
While a lot of attempts were made, a major breakthrough to the dynamic, legged locomotion happened in early 1980s at Caltech with the developments of the group of Mark Raibert \cite{raibert1986legged}.
They have engineered a one-leg robot that was built purposely to perform fast dynamical movements rather than to resemble a humanoid or quadruped.
With these developments, it has become possible to pronk, jump and make backflips.

\begin{figure}[hbt]
    \centering
    \includegraphics[width=1\linewidth]{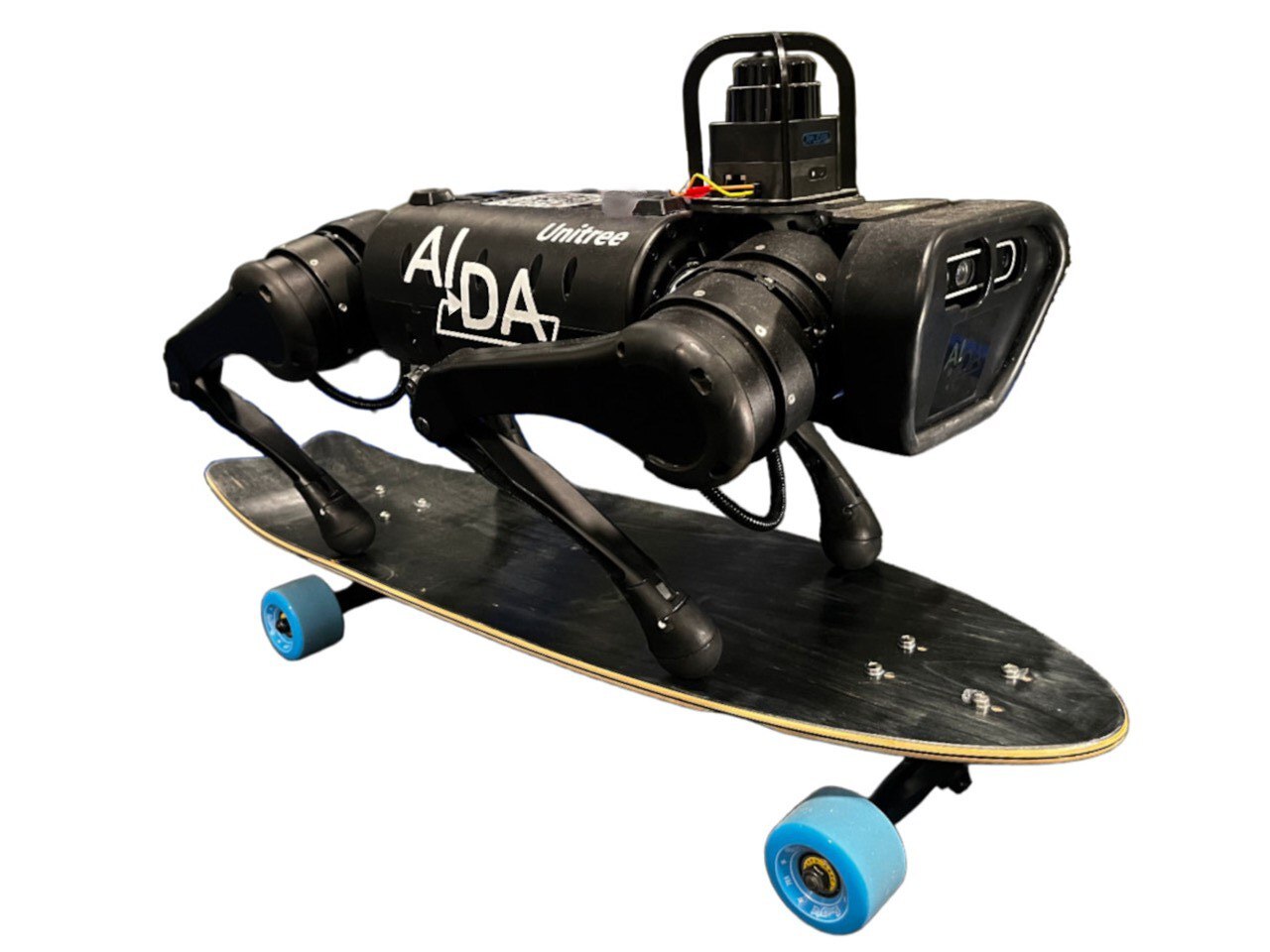}
    \caption{Unitree A1 quadrupedal robot on a skateboard. Autonomous mounting, i.e. climbing on the board, is a challenging problem, and was not yet considered.}
    \label{fig:1.jpg}
\end{figure}

As impressive as it is, those robots were externally powered, they had external control, they were expensive and extremely difficult to operate.
It took long time for these control approaches and this type of machines to reach the market.
One of the most evident obstacles to that were the actuators.
Hydraulics is powerful, but difficult to maintain, develop and operate.

An alternative to the hydraulic actuation is the electric one.
Electric motors, while having considerably smaller power density, are easier to work with and to build a robot upon.
However, in order to design a mobile robot with electric actuation, some of the desired characteristics had to be sacrificed.
High output torque was provided with huge gearboxes, leading to slow acceleration.

It changed with the introduction of the motors like the ones used in MIT Cheetah \cite{seok2012actuator} and much cheaper version that inherited its advantages \cite{katz2018low}.
With the introduction of these machines and their descendants to the market it has become possible to apply modern control methods, such as Model Predictive Control (MPC) to the torque-controller robots in dynamic scenarios.
A number of works were published shortly after, exploring the capabilities of quadrupeds in various scenarios, like jumping, backflipping, running, climbing obstacles, traversing challenging outdoor terrains in dynamic fashion.

One of the remaining caveats of the motors used in modern quadrupedal robots is their power consumption.
Due to the high conductivity of the coils and high currents, the heat losses can be as high as 76$\%$ \cite{kau2019stanford}.

In order to improve the efficiency of quadrupedal robots, a number of works were devoted to supplementing quadrupeds with wheels.
Overall, wheeled platforms are much simpler mechanically.
They rely on the continuous contact point change, allowing for more straightforward control algorithms.
A clear advantage of the wheeled platforms unfolds in the locomotion on even terrains.

There are two ways to make quadrupedal robot move on wheels: they can be active and passive.
Using active wheels streamlines the development of the control algorithms, as they do not introduce any new degrees of underactuation.
Non-actuated wheels require the passive dynamics of the legged-wheeled platform to be considered, and in the case of a normal skateboard additional constraints should be enforced to prevent slippage of the feet from the board.
The riding of a skateboard by a quadruped was recently presented, and this work is devoted to the extension of the capabilities of the quadrupedal robots to ride skateboards autonomously (\autoref{fig:1.jpg}).

\section{Related Work}

Legged robots have demonstrated remarkable progress in recent years, achieving versatile locomotion across diverse terrains.
An emerging frontier in legged robotics research involves interaction with everyday objects, particularly skateboards, which presents unique challenges in dynamic balance and control.

Several recent works have made significant contributions to robot skateboarding.
Kim et al. \cite{kim2021bipedal} developed LEONARDO, a bipedal robot capable of performing skateboarding maneuvers using a combination of legs and propellers.
Thibault et al. \cite{thibault2024learning} demonstrated skateboarding for humanoid robots through massively parallel reinforcement learning, achieving an efficient pushing motion with the REEM-C platform.
Liu et al. \cite{liu2025discrete} introduced Discrete-time Hybrid Automata Learning (DHAL), enabling quadrupedal robots to perform skateboarding through on-policy reinforcement learning that identifies and executes appropriate mode-switching behaviors.
Their approach used a beta policy distribution with a multi-critic architecture to effectively handle contact-guided motions.
Complementing these learning-based approaches, Xu et al. \cite{xu2024optimization} developed an optimization-based control pipeline for the CyberDog2 platform to achieve dynamic balance and acceleration on a skateboard.
Earlier work by Bjelonic et al. \cite{bjelonic2018skating} explored related hybrid mobility with a force-controlled quadrupedal robot performing skating motions using passive wheels and ice skates, demonstrating significant energy efficiency improvements over traditional walking gaits. Similarly, Chen et al. \cite{chen2020roller} investigated dynamic roller-skating for mammalian quadrupedal robots with passive wheels, proposing a pace-like gait that achieved higher speed and energy efficiency compared to conventional trotting, while maintaining dynamic stability through bilateral symmetry.

There is a fundamental limitation across all these approaches: they begin with the robot already positioned on the skateboard, addressing only the riding phase of the interaction.
The critical challenge of autonomously mounting a skateboard — a complex, contact-rich maneuver requiring precise coordination and dynamic stability — remains unexplored in the literature.
This mounting phase presents unique difficulties that existing approaches do not address, representing a significant gap between current capabilities and fully autonomous skateboarding robots.

The skateboard mounting task is particularly challenging for several reasons.
First, it presents a sparse reward problem where successful completion requires a precisely executed sequence of movements.
Traditional reinforcement learning approaches struggle with such problems, as random exploration is unlikely to discover successful strategies.
Second, the transition from quadrupedal stance to skateboard mounting demands careful coordination of multiple limbs while maintaining balance throughout the motion.
Finally, mounting a potentially moving skateboard introduces additional complexity, requiring the robot to adapt to the board's dynamics while executing precise foot placements.

In this paper, this critical research gap is addressed by a goal-oriented reverse curriculum learning approach that enables a quadrupedal robot to autonomously mount and ride a skateboard.
In the proposed approach the learning begins from the successfully mounted states and gradually expands backward to include more challenging initial conditions, addressing the sparse nature of the reward in this problem.

\section{Control Algorithm}

While choosing a control paradigm to employ, two families of approaches were considered, Model Predictive Control (MPC) and Reinforcement Learning (RL).

MPC is one of the most widely adopted methods for controlling quadrupedal robots due to its optimization-based ability to plan motions while taking into account the system dynamics and constraints.
It is particularly effective for complex dynamic tasks such as running, gait transitions, and jumping, as it allows the computation of ground reaction forces and Center of Mass (CoM) trajectories while accounting for the contact sequences and body inertia \cite{ding2021representation}.

Despite their effectiveness, these methods require predefined motion plans and rely heavily on the reliability of the data that the perception modules provide\cite{manchester2019variational}.
This dependency becomes a critical drawback in complex, unknown, or uncertain environments.
Changes in terrain, such as the displacement of a structure that the robot intends to climb, invalidate the pre-computed motion plans.
Consequently, the feasible trajectories should be re-generated in real time, significantly increasing the computational demands to the on-board computer.

Under the circumstances of a complex environment, that is difficult to observe and model precisely, another path can be taken.
In contrast to MPC, RL algorithms optimize for the parameters of the controller, that itself generates the control during the inference.

Prior work has shown that RL enables robust and agile behaviors in similar scenarios, such as fast locomotion \cite{rapidLocomotion}, safe high-speed navigation \cite{rlNavigation}, and terrain-adaptive movement with limited perception \cite{limPerception}, as well as manipulation of dynamic objects \cite{leggedManipulation}. These capabilities are especially relevant when interaction with everyday objects, like skateboards, requires whole-body coordination and flexible adaptation \cite{Parkour}.

Among RL algorithms, Proximal Policy Optimization (PPO) is widely used for continuous control tasks due to its simplicity, stability \cite{stability1, stability2}, and strong empirical performance. PPO-based architectures have been successfully used in modular or hierarchical setups for skill switching \cite{HierarchicalRL, RLforversatile, robustrec}, and they are particularly effective when combined with curriculum learning in sparse-reward environments. 

We used the implementation of PPO from the original Isaac Lab environment, owing to its demonstrated strong performance in locomotion tasks both on flat and, critically, on rough terrain, which exhibits similarities to the task at hand.
Furthermore, the core model architecture, a multilayer perceptron (MLP) with three hidden layers and ELU activation functions, was preserved.

In this work, PPO was integrated with a task-specific curriculum to enable a quadrupedal robot to learn skateboard climbing. This formulation was inspired by recent work on contact-rich control through reinforcement learning with hybrid dynamics \cite{JMLRv13ly12a}, where a discrete-time hybrid automaton is jointly trained with the policy to explicitly learn contact-driven mode switching and continuous dynamics within each mode \cite{liu2025discrete}. This framework achieves stable skateboarding without requiring trajectory segmentation or mode labels, and demonstrates strong sim-to-real transfer capabilities in a highly dynamic underactuated setting.

To address the challenge of initializing training in environments with low success probability, Reverse Curriculum Learning method proposes to start learning from states near the goal and progressively move towards more distant initial conditions. This strategy aligns with the structure of skateboard mounting, where any prior  knowledge other than a single state in which the task is achieved is not needed \cite{ReverseCurriculum}. The method used in this work builds upon this foundation by integrating reverse curriculum into the broader RL pipeline to overcome sparse-reward approach.


\section{Experimental Design}

\subsection{Environment}
The experiments utilize the RobotLab-Isaac-Velocity-Flat-Unitree-A1-v0 environment, developed within the robot\_lab extension \cite{fan2024}. The environment is built on the Isaac Lab framework \cite{mittal2023orbit} and is designed for robotic simulation and learning.

To enable our research, we extended this environment by designing and integrating a physically realistic skateboard model. It features accurate physical properties, with a weight of 2.1 kg and deck dimensions of 575 × 250 mm. To realistically simulate the dynamics of truck suspension, we modeled the mechanism using a P controller for position control, effectively replicating the behavior of a torsion spring in the truck joint. 

\subsection{Mathematical Model of the Skateboard}

The skateboard truck mechanism converts the deck tilt into wheel turning, formalized as:

\begin{equation}
\beta \approx \alpha
\end{equation}

where $\alpha$ is the deck tilt angle and $\beta$ is the truck turning angle, with this 1:1 relationship resulting from the $45^{\circ}$ kingpin angle.

The bushings in the truck act as torsional springs, modeled as:

\begin{equation}
\tau_{bushing} = -k_{angular} \cdot \beta - c \cdot \dot{\beta}
\end{equation}

where $k_{angular}$ is the angular stiffness coefficient (2 N$\cdot$m/rad in our medium trucks) and $c$ is the damping coefficient. In the Isaac simulation, this mechanism was implemented using a P controller to replicate the torsion spring behavior.

The dynamics of the skateboard follow rigid body equations:

\begin{equation}
m\dot{\mathbf{v}} + \boldsymbol{\omega} \times (m\mathbf{v}) = \mathbf{F}_{ext}
\end{equation}

\begin{equation}
\mathbf{I}\dot{\boldsymbol{\omega}} + \boldsymbol{\omega} \times (\mathbf{I}\boldsymbol{\omega}) = \boldsymbol{\tau}_{ext}
\end{equation}

For wheel-ground contact, we enforce a no-slip constraint:

\begin{equation}
\text{tg}\left(\frac{f_x}{f_z}\right) \leq \mu
\end{equation}

where $f_x$ and $f_z$ are the horizontal and vertical contact force components, and $\mu$ is the friction coefficient. Similarly, for robot-skateboard interaction:

\begin{equation}
|\mathbf{f}_{tangential}| \leq \mu_{robot-skateboard} \cdot |\mathbf{f}_{normal}|
\end{equation}
Throughout most of the training process, the skateboard was fixed in position, allowing the robot to focus on learning the mounting behavior. The skateboard was allowed to move freely only in the final stage of training, enabling the robot to adapt to the dynamic nature of an unfixed board.

\subsection{Observation and Action Space}

The base environment has the following configuration:

\begin{itemize}
  \item Observation space:
    \begin{itemize}
      \item The positions and velocities of all joints, represented as vectors $\mathbf{q} \in \mathbb{R}^{12}$ and $\dot{\mathbf{q}} \in \mathbb{R}^{12}$ respectively.
      \item The angular velocity of the robot base, represented as a vector $\boldsymbol{\omega} \in \mathbb{R}^{3}$.
      \item The projected gravity vector, $\mathbf{g} \in \mathbb{R}^{3}$, to provide information regarding the robot's orientation.
      \item The linear velocity of the base frame, $\mathbf{v} \in \mathbb{R}^{3}$ (available only for the critic network).
    \end{itemize}
  \item Action space:
      \begin{itemize}
        \item Target positions of 12 joints, $\mathbf{a} \in \mathbb{R}^{12}$.
      \end{itemize}
\end{itemize}

To adapt the environment to the task requirements, the observation space was augmented with information pertaining to the skateboard.  This includes the position and orientation of the skateboard, represented within the robot’s coordinate frame. We also add positions of points along the skateboard’s edge, spaced about 10 cm apart, to represent the board’s shape. This allows the robot to work with different skateboards, even if we don’t have exact models. A 10 cm interval gives a good balance between precision and computational efficiency, but should be fine-tuned. This extra information can be obtained from a depth camera, facilitating the transfer of learned policies from simulation to a physical robot platform. Given the critical influence of the robot’s foot positions and contact status with the skateboard on task performance, these variables were also integrated into the observation space. However, acknowledging the inherent difficulty in distinguishing skateboard contact from contact with other environmental elements in a real-world setting, it is planned to exclude this component in subsequent investigations.

\subsection{Reward Function}

Throughout the entirety of the training procedure, all default reward functions inherent to the simulation environment remained active. Table~\ref{tab:reward_terms_extended} details the supplementary reward terms that were introduced to facilitate the skateboard mounting task. Sparse, contact-based rewards were employed to incentivize contact between each foot and the upper surface of the skateboard deck. In addition, dense, distance-based rewards, modulated by an exponential kernel, were utilized to encourage the alignment of the robot’s and skateboard’s centers of mass (COMs) and coordinate axes.  

\begin{table}[t]
\caption{Summary of Reward Terms and Their Expressions.}
\centering
\renewcommand{\arraystretch}{1.5}
\begin{tabular}{ll}
\hline
\textbf{Mounting Reward} & \textbf{Expression}\\
\hline
Feet on board & $\ \sum_{i=1}^4 \mathbf{1}_{\text{contact}, i}$ \\
Orientation alignment & $\ \exp\left(-\frac{1}{\sigma^2} \cdot \frac{\theta_{\text{rel}}}{\pi} \right) \cdot \mathbf{1}_{d < d_{\text{th}}}$ \\
Distance to skateboard & $\ \exp\left(-\frac{d}{\sigma^2} \right)$ \\
Skateboard flip & $\ \mathbf{1}_{g_z > 0}$ \\
Skate velocity penalty & $\ \|\mathbf{v}_{\text{skate}}\|$ \\
\hline
\end{tabular}
\label{tab:reward_terms_extended}
\end{table}

\textbf{Notation:} $\mathbf{1}_{\text{contact}, i}$ - indicator function for contact of foot $i$ with the skateboard; $\theta_{\text{rel}}$ - relative yaw angle between the robot and skateboard frames; $d$ - distance between the robot's base and the skateboard; $d_{\text{th}}$ - distance threshold for considering orientation alignment reward; $g_z$ - $z$-component of the projected gravity vector in the skateboard frame; $\mathbf{v}_{\text{skate}}$ - linear velocity of the skateboard in the horizontal plane; $\sigma$ - scaling factor for spatial distance-based terms.

\subsection{Baseline Training}

The training process commenced with a gait training task, pre-implemented within the simulation environment, to develop fundamental locomotion skills. The robot learned a stable gait pattern, guided by corresponding reward functions, while concurrently tracking velocity commands issued by the command manager. During this initial phase, only the default reward functions inherent to the environment were utilized. Upon achieving a stable gait, the supplementary reward terms, as delineated in the preceding section, were introduced, and novel velocity commands were generated to direct the robot toward the skateboard.

As a baseline comparison, a forward curriculum approach was implemented wherein the robot was spawned in the vicinity of the skateboard, with initial position and orientation sampled uniformly. The most promising result observed with this method was successful alignment of the robot’s center of mass and coordinate axes with those of the skateboard, along with the placement of two legs on the board. A series of reward shaping strategies were subsequently investigated to enhance performance, including an exponential reward function for skateboard contact, supplementary rewards for individual foot approaching near target locations, and penalties for link deviation from desired positions. Nevertheless, none of these modifications led to a significant improvement in the final outcome.

\subsection{Reverse curriculum learning}

\begin{figure*}[hbt]
    \centering
    \includegraphics[width=\textwidth]{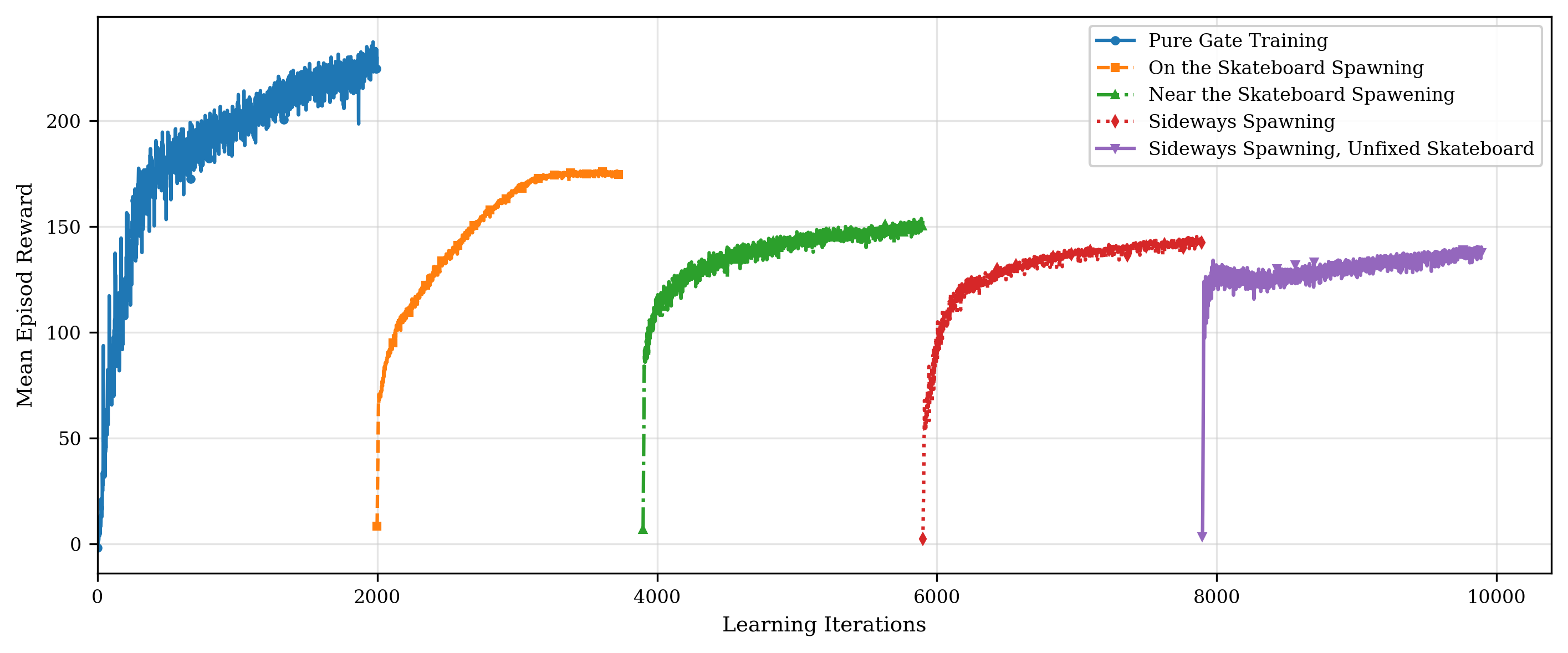}
    \caption{Training performance curves across different curriculum stages for the quadruped skateboard mounting task, showing mean episode reward versus learning iterations}
    \label{fig:2.png}
\end{figure*}

In many goal-oriented reinforcement learning problems, reaching the goal from most initial states requires impractical amounts of exploration. However, agents find it much easier to reach goals from nearby states due to the strong, readily available reward signal. After learning to reach goals from these proximal states, the agent can then train from more distant states by bootstrapping from its existing knowledge.

\begin{table}[ht]
\centering
\caption{Network Architecture and Training Hyper-parameters}
\renewcommand{\arraystretch}{1.5} 
\begin{tabular}{p{0.3\textwidth} p{0.11\textwidth}}
\hline
\textbf{Network Hyperparameters} & \textbf{Value} \\
\hline
Actor Hidden Dims & [1024, 512, 256] \\
Critic Hidden Dims & [1024, 512, 256] \\
\hline
\textbf{PPO Hyperparameters} & \textbf{Weight} \\
\hline
Value Loss Coefficient & 1.0 \\
Use Clipped Value Loss & True \\
Clip Parameter & 0.2 \\
Entropy Coefficient & 0.01 \\
Number of Learning Epochs & 5 \\
Number of Mini-Batches & 4 \\
Learning Rate & 1.0e-3 \\
Schedule & adaptive \\
Gamma & 0.99 \\
Lambda & 0.95 \\
Desired KL Divergence & 0.01 \\
Max Gradient Norm & 1.0 \\
\hline
\textbf{Training parameters} & \textbf{Weight} \\
\hline
Total Number of Episodes & 10000 \\
Length Of The Episode, s & 5 \\
Environments & 4096 \\
Training Time, hours & 7 \\
\hline
\end{tabular}
\label{tab:implementation_details}
\end{table}

Following this principle, the robot was spawned directly above the skateboard. The range of initial joint positions and velocities was also substantially narrowed, and external random forces and torques, which were employed during the initial training stage, were eliminated. It has been experimentally determined that expanding the distribution of initial states during this training phase prevents the agent from effectively learning the desired policy. Instead of achieving the intended behavior, the agent frequently converges towards a state where it places a single foot on the ground, a configuration that appears to offer greater stability despite being inconsistent with the target objective. During this phase, the agent learned to maintain its position at the target point, a challenging task in itself due to the skateboard’s limited surface area.

After the robot learned how to balance on a skateboard, we moved on to the next stage of training. At this stage, the spawn point was selected evenly inside a square with a side of 60 by 60 cm with the center coinciding with the center of mass of the skateboard. The range of initial joint positions, velocities and external random forces and torques has also been increased. The weights of the awards and all other training parameters have not changed.

Following the preceding stage, the robot successfully learned to mount the skateboard, initiating movement from a random point within its immediate vicinity. However, further expansion of the spawning region proved infeasible, as the robot’s gait quality degraded significantly during the training process. Consequently, it was concluded that simultaneously addressing the locomotion and skateboard-mounting tasks within a single neural network was not expedient. Therefore, in the subsequent training stage, the robot’s initial position was set adjacent to the skateboard, within a region that could be reached utilizing the neural network trained in the first stage.

\begin{figure*}[ht]
    \centering
    \begin{tabular}{@{}ccc@{}}
        \includegraphics[width=0.31\textwidth]{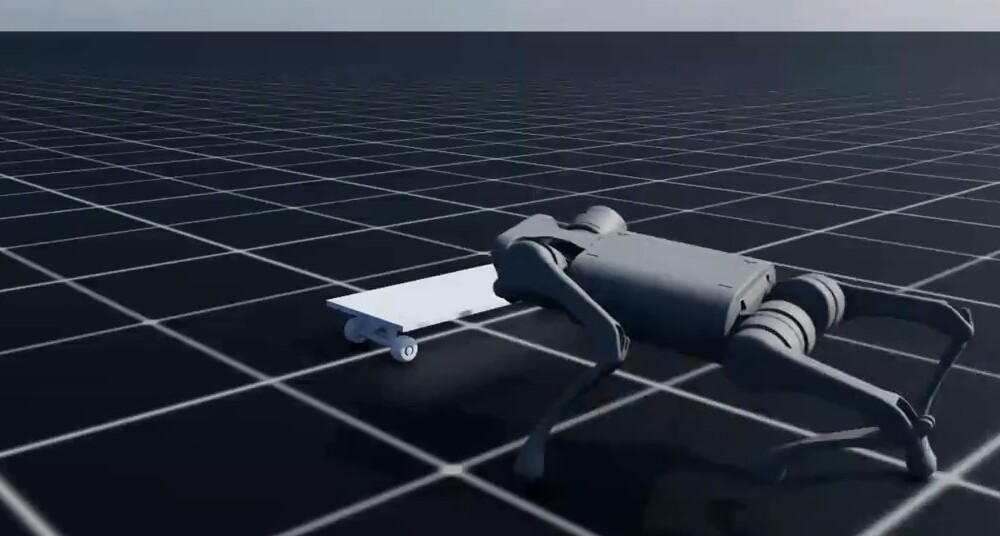} &
        \includegraphics[width=0.31\textwidth]{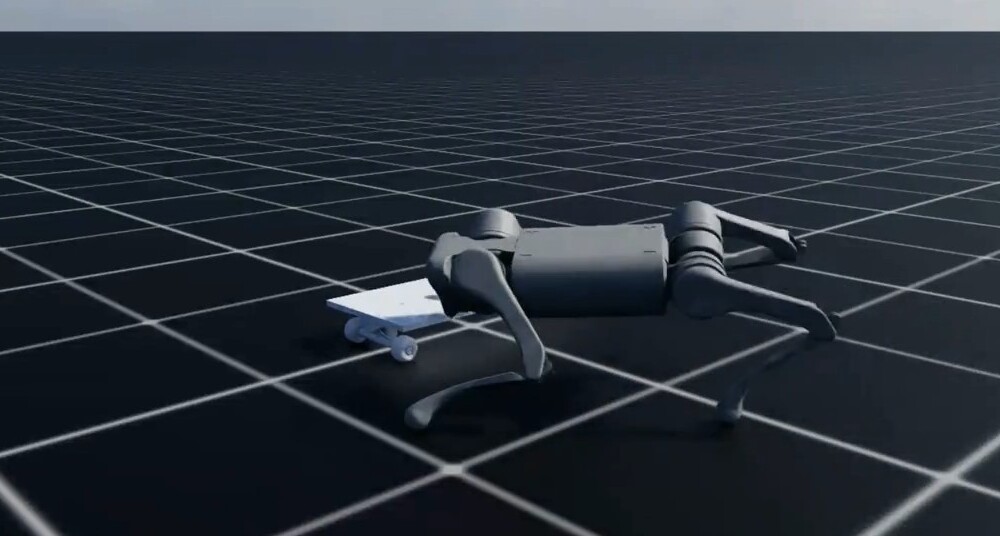} &
        \includegraphics[width=0.31\textwidth]{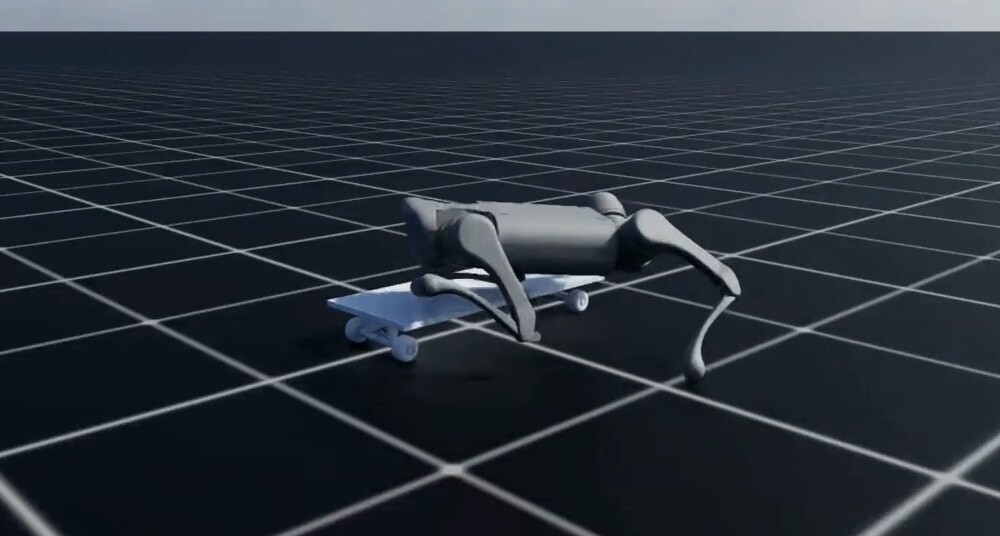} \\
        \footnotesize (a) Initial approach & 
        \footnotesize (b) First contact & 
        \footnotesize (c) Transition phase \\[0.5cm]
        
        \includegraphics[width=0.31\textwidth]{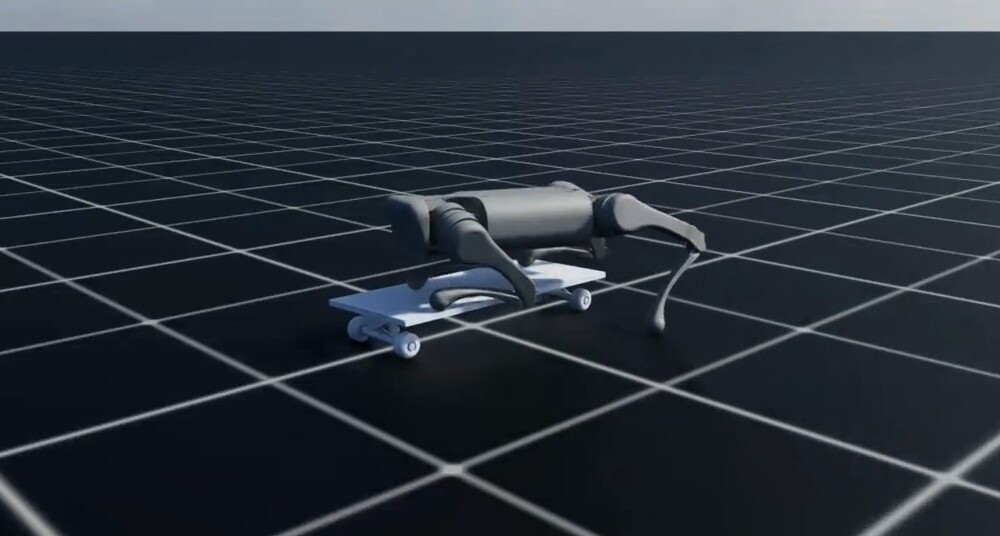} &
        \includegraphics[width=0.31\textwidth]{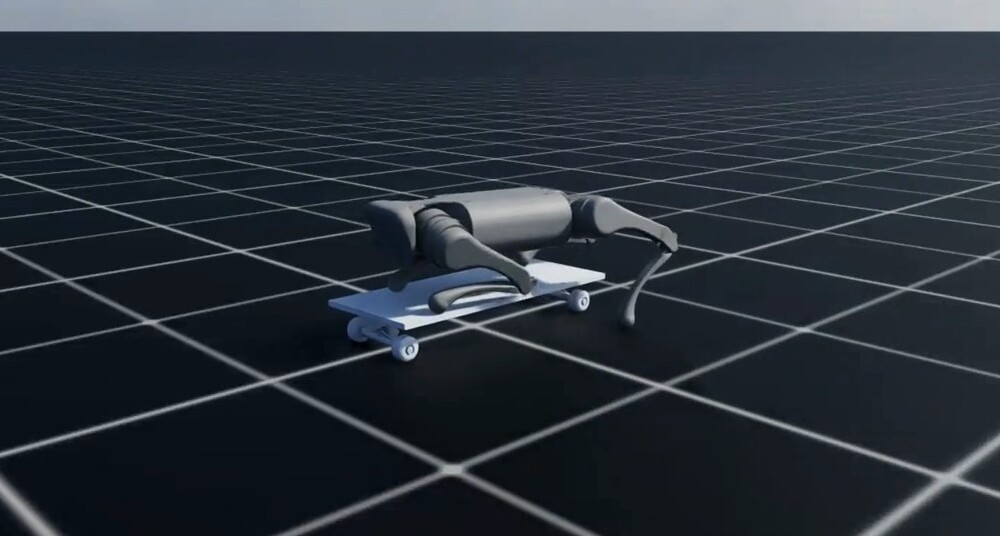} &
        \includegraphics[width=0.31\textwidth]{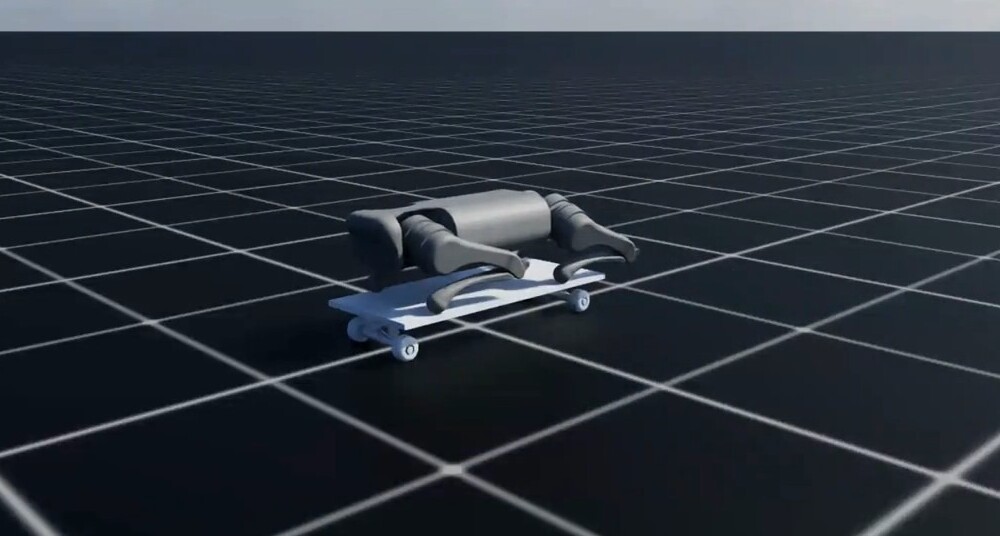} \\
        \footnotesize (d) Progress phase & 
        \footnotesize (e) Near completion & 
        \footnotesize (f) Mounted state
    \end{tabular}
    \caption{Sequence of the quadrupedal robot skateboard mounting process. Unitree A1 progresses from (a) initial approach to (f) fully mounted state.}
    \label{fig:mounting_sequence}
\end{figure*}

In the final stage, we investigated the scenario involving an unfixed skateboard. Notably, the agent trained on a fixed skateboard demonstrated good performance without requiring additional training. To further refine the results, an episode termination condition was introduced, triggered by the skateboard overturning, along with negative rewards for both overturning events and skateboard displacement. The agent’s initial state was consistent with that employed in the preceding training stage.

 \autoref{fig:2.png} shows mean episode reward curves for all stages of training process.

\subsection{Implementation Details}

In Table~\ref{tab:implementation_details}, we outline the network parameters, hyperparameters for PPO and details of the training process. A single NVIDIA RTX4090 GPU with 24GB of VRAM was used for training.

\subsection{Results}

The trained policy successfully achieved skateboard mounting from various starting positions. \autoref{fig:mounting_sequence} illustrates the complete mounting sequence. Our approach demonstrates robust performance, with the robot consistently mounting the skateboard in approximately 3 seconds from first contact.

Additional demonstration and results are available in the project repository.

\section{Conclusion and Future Work}

A method was proposed for the Reverse Curriculum Reinforcement Learning-based mounting of a skateboard in simulation by a quadrupedal robot.
It relies on the gradual increase of the problem that is being solved, from trying not to fall from the board to full mounting, when the robot is initialized far from the skateboard.
Numerical experiments were conducted in simulation.

The results suggest that robust skateboard mounting can be performed by a quadruped in short time.
Overall, this work addresses the gap between walking and riding in quadrupedal robots towards creating a complete skill pipeline for autonomous skateboarding — from approach to mounting to riding.

Future work may encompass the following:
\begin{itemize}
    \item Combining the mounting skill and the riding skill in one controller
    \item Validating the proposed approach with the real-world experiments
    \item Introducing the steering control via leaning the body of the robot to the sides
\end{itemize}

\bibliographystyle{IEEEtran}
\bibliography{root}

\end{document}